\documentclass{bmcart}

\RequirePackage{natbib}
\usepackage[utf8]{inputenc} 
\usepackage{graphicx} 
 \usepackage[caption=false,font=footnotesize]{subfig}
\usepackage[cmex10]{amsmath} 
\usepackage{amssymb}  
\usepackage{color}
\usepackage{bbm}
\usepackage{algorithm,algorithmicx,algpseudocode}
\usepackage{url}
\usepackage{float}
\usepackage{eurosym}
\usepackage{multirow}
\usepackage{soul}
\usepackage{ragged2e}

\newcommand{\amsterdam}{Amsterdam}
\newcommand{\berlin}{Berlin}
\newcommand{\firenze}{Florence}
\newcommand{\copenhagen}{Copenhagen}
\newcommand{\milano}{Milan}
\newcommand{\muenchen}{Munich}
\newcommand{\roma}{Rome}
\newcommand{\stockholm}{Stockholm}
\newcommand{\torino}{Turin}
\newcommand{\wien}{Vienna}

\newif\iffigures
\figurestrue 

\iffigures
\else
	
	\def\includegraphics{}
\fi

\startlocaldefs
\endlocaldefs

\begin{document}

\begin{frontmatter}

\begin{fmbox}
\dochead{Research}


\title{Weak Signals in the Mobility Landscape:\\ Car Sharing in Ten European Cities}


\author[
   addressref={aff1},                   
   corref={aff1},                       
   email={chiara.boldrini@iit.cnr.it}   
]{\inits{CB}\fnm{Chiara} \snm{Boldrini}}
\author[
   addressref={aff1},
   email={raffaele.bruno@iit.cnr.it}
]{\inits{RB}\fnm{Raffaele} \snm{Bruno}}
\author[
   addressref={aff1},
   email={haitam.laarabi@iit.cnr.it}
]{\inits{HL}\fnm{Haitam} \snm{Laarabi}}


\address[id=aff1]{
  \orgname{IIT-CNR}, 
  \street{Via G. Moruzzi 1},                     %
  \city{Pisa},                              
  \cny{Italy}                                    
}


\begin{artnotes}
\end{artnotes}

\end{fmbox}


\begin{abstractbox}

\begin{abstract} 
\justify
Car sharing is one the pillars of a smart transportation infrastructure, as it is expected to reduce traffic congestion, parking demands and pollution in our cities. From the point of view of demand modelling, car sharing is a weak signal in the city landscape:  only a small percentage of the population uses it, and thus it is difficult to study reliably with traditional techniques such as households travel diaries. In this work, we depart from these traditional approaches and we leverage web-based, digital records about vehicle availability in 10 European cities for one of the major active car sharing operators. We discuss which sociodemographic and urban activity indicators are associated with variations in car sharing demand, which forecasting approach (among the most popular in the related literature) is better suited to predict pickup and drop-off events, and how the spatio-temporal information about vehicle availability can be used to infer how different zones in a city are used by customers. We conclude the paper by presenting a direct application of the analysis of the dataset, aimed at identifying where to locate maintenance facilities within the car sharing operation area. 
%
\end{abstract}


\begin{keyword}
\kwd{car sharing}
\kwd{smart transportation}
\kwd{urban computing}
\kwd{data mining}
\end{keyword}


\end{abstractbox}
%

\end{frontmatter}



\section{Introduction}
\label{sec:intro}
\noindent
%
Automobile transportation has been one of the main drivers of the population growth and increasing wealth that have characterised the last two centuries~\cite{mitchell2010reinventing}. Thanks to cars, people have had greater access to jobs, goods, services. However, these benefits have not come for free. The price paid for our increased mobility has been huge in terms of environmental pollution, city congestion and resulting health issues. We are now at a turning point for personal mobility systems: policy makers and citizens share the common idea that it is time to rethink the way we move. There are three main driving forces behind this personal mobility revolution: smart transportation, sharing economy, and green vehicles, all tightly intertwined. The departure from ownership mindset to usage mindset will make it possible to have significantly fewer vehicles in our cities. The implications are that we can save space (public parking space and private garage space) and use it for something with increased added value than to host idle cars for hours (a private car is used only $5\%$ of its available time, corresponding to 72 minutes per 24 hrs~\cite{parkingtime}). This usage mindset will also allow people to rent the car size most appropriate to their daily needs, thus implementing the Mobility-as-a-Service concept. Since the average vehicle has only around $1.5$ occupants~\cite{occupancyus,occupancyeu}, people can refrain from buying a car able to address the extreme case of personal mobility (e.g., moving a whole family for a vacation) and instead use two-seaters, which are more suitable for everyday commuting. On occasion, they will be able to rent larger vehicles if needed. The virtuous mobility cycle is completed with the switch to electric vehicles, which allow for a drastic reduction in the carbon footprint of personal mobility.

In this framework, car sharing is emerging as one the most promising examples of Mobility-as-a-Service~\cite{shaheen2015mobility}. The general idea of car sharing is that the members of a car sharing system can pick up a shared vehicle of the car sharing fleet when they need it. Different operators may implement different pickup/drop-off policies. In station-based systems, members can only pick up and drop off vehicles at designated locations called stations, as in the Autolib system in Paris. If the service is two-way (e.g., Zipcar, Modo), people are asked to bring back the vehicle to the station where they initially picked it up. Otherwise, the service is called one-way. One-way services are definitely the most popular among customers thanks to the flexibility they provide. Examples of one-way car sharing are Autolib, Ha:Mo ride, CITIZ. One-way services can drop altogether the concept of station: this is the case of so-called free floating car sharing -- such as Car2go, DriveNow, Enjoy -- whose customers can pick up and drop off vehicles anywhere within a predefined operation area.

Car sharing is a \emph{weak signal}  in the city landscape: the fraction of people relying on car sharing for their daily trips is rapidly increasing but it is still in the order of single digit percentage points in the best cases~\cite{kortum2014driving}. So far, car sharing has been mostly studied through surveys and direct interviews with its members~\cite{schwieger2015global,shaheen2015mobility}. In addition, car sharing is typically not accounted for in households travel diaries periodically collected by city administrations. Even if it were, the limitations of travel surveys are widely acknowledged, and range from their inability to capture changes in the routine travel behaviour to their underestimation (because of underreporting from people) of short, non-commute trips~\cite{STOPHER2007}. Moreover, running a survey is very expensive if one wants to capture a statistically meaningful sample. 

\begin{table}[!t]
\caption{General information on the 10 cities (area: $km^{2}$, population density: people/$km^{2}$; education: persons aged 25-64 with ISCED level 5, 6, 7 or 8 as the highest level of education; mean daily temperature: minimum/maximum average daily temperature over the year). Source: Urban Audit Database~\cite{urbanaudit} \& Wikipedia.}
\centering
\begin{tabular}{l|r|r|r|r|r|r}
City & {\euro}GDP/capita & Population & Area & Pop. Density & Education & Mean daily $C^\circ$\\
\hline \hline
\amsterdam & 46,952 & 853,312 & 165.76 & 5147.876 & 254,000 & 3.4/17.6\\
\hline
\berlin & 35,627 & 3,520,031 & 891.68 & 3947.639 & 752,300 & 0.6/19.2\\
\hline
\firenze & 31,547 & 382,929 & 102.32 & 3742.465 & 59,627 & 6.5/24.6\\
\hline
\copenhagen & 70,183 & 559,440 & 88.25 & 6339.263 & 152,817 & 0/17\\
\hline
\milano & 87,786 & 1,368,590 & 181.67 & 7533.380 & 224,256 & 2.5/23.6\\
\hline
\muenchen & 46,377 & 1,450,381 & 310.70 & 4668.107 & 371,200 & 0.3/19.4\\
\hline
\roma & 55,385 & 2,874,529 & 1287.36 & 2232.887 & 415,766 & 7.5/24.5\\
\hline
\stockholm & 81,395 & 939,238 & 187.16 & 5018.369 & 234,787 & -1.7/18.8\\
\hline
\torino & 74,725 & 886,837 & 130.17 & 6812.910 & 109,314 & 2.9/20.8\\
\hline
\wien & 58,140 & 1,867,582 & 414.87 & 4501.608 & 232,009 & 1.2/21.7\\
\end{tabular}

\label{tab:cities_general_info}
\end{table}%

Cities have been considered kaleidoscopes of information since a long time~\cite{meier1962communications} but the extent to which this is true has reached new heights now that a myriad of electronic devices have weaved into its fabric. From the car sharing perspective, this means that we can now know exactly when and where cars are available, and we can observe shared vehicle flows \emph{as they happen} in the city. This knowledge opens up a new avenue of research that goes in the direction of the new science of cities and urban computing: using data and electronic devices to extract knowledge and to improve urban solutions. Along these lines, the goal of this paper is to stimulate a discussion on how to apply urban computing ideas to the car sharing domain. To this aim, we exploit the availability of public, web-based data about free floating car sharing in 10 European cities (whose main characteristics are summarised in Table~\ref{tab:cities_general_info} for the convenience of the reader) and we carry out an analysis with the following objective in mind: to understand what mining this kind of data can bring to cities and to car sharing operators alike. The main contributions of this study can be summarised as follows:

\begin{itemize}
\item We perform an explanatory analysis of the car sharing demand as a function of the sociodemographic and urban fabric (i.e., number, heterogeneity, and category of Foursquare Points of Interests - PoI) indicators associated with the cities of \milano{}, \roma{}, and \torino{}\footnote{Only for these cities we were able to find fine-grained geospatial census data significantly overlapping with the car sharing operation area.}. While a single explanatory pattern does not emerge across the cities, they share indeed several similarities. In fact, their car sharing demand is positively associated with high educational attainment (all Italian cities) and negatively correlated with commuting outside of the municipality area (\milano{}, \roma{}). \textcolor{black}{These findings confirm the conclusion of the most recent sociodemographic surveys about car sharing services~\cite{Kopp2015,Wittwer2018,Giesel2016,Becker2017}, but at a much finer spatial granularity and without relying on expensive and time-consuming interviews/questionnaires. With regards to the urban fabric indicators, the only PoI category that seems to have a statistically significant effect on car sharing demand is that of nightlife-related activities, suggesting that leisure is the most typical trip purpose.} 
\item \textcolor{black}{We take into consideration several approaches to demand forecasting, and we evaluate which are the best performing when it comes to car sharing pickups/drop-offs forecasting. Our results show that Random Forest yields consistently better results than simple average-based forecasting, time series forecasting, vanilla neural networks, and a popular custom approach proposed in the literature. However, prediction quality is in general quite good, even with the simplest solutions. } 

\item Four distinct car availability temporal patterns can be recognised in the cities considered in this study. We have labelled them \emph{day}, \emph{night}, \emph{neutral}, and \emph{high-intensity} behaviours, based on when they exhibit their peak availability and on the intensity of this peak. We also show that these patterns tend to be spatially autocorrelated, i.e., neighbouring cells are likely to feature the same behaviour. 
\item Motivated by the importance that customers place on the cleanliness of vehicles, we propose a simple approach to the effective deployment of car sharing maintenance facilities. We show that including the airport zone in the operation area and locating maintenance facilities there is a simple yet effective strategy to reduce the maintenance trips carried out by the car sharing workforce.
\end{itemize}
\color{black}

\section{Related work}
\label{sec:relwork}

In the following we provide a brief overview of the most relevant works in the area of data science for car sharing, data science for transportation systems in general, and data-driven car sharing operation models. 

\color{black}

\subsection{Knowledge mining from survey data}
\label{sec:survey}

Until recently, knowledge about car sharing systems has been mostly acquired through surveys, in which car sharing operators and members are interviewed. The main goal of these studies is to characterise the sociodemographic profile of car sharing users, as well as investigating the reasons behind their choices and the impact that car sharing has had on their mobility behaviour. In 2005, Millard-Ball~\cite{Millard-Ball2005} presented one of the first comprehensive sociodemographic  analysis of station-based car sharing in North America, highlighting a few key demographics indicators that will constantly reappear also in analyses of more recent car sharing solutions. After interviewing 978 US and 362 Canadian car sharing members through a web-based survey, Millard-Ball reports that car sharing members are typically young (25-44 year old), with high income and well-educated. They live in small households, often with no private cars. This survey does not support the finding, often presented in the related literature, that car sharing members are typically male. Recreational trips, shopping-related trips, and personal business trips are by far the most popular trip purpose for the respondents. In 2010, these findings are substantially confirmed by~\cite{Momo2010} for Europe, with the interesting addendum that car sharing customers tend to have season tickets for public transport more than the general population. 

Considering that free floating car sharing is a recent addition to the car sharing domain (e.g. Car2go was founded in 2008, and started a significant expansion only in 2011), in the following we overview recent surveys~\cite{Kopp2015,Wittwer2018,Giesel2016,Becker2017} focusing specifically on the free floating modality. Kopp at al.~\cite{Kopp2015} recruited 204 males between 25 and 45 years of age living in the cities of Munich and Berlin, Germany. 109 were free floating car sharing members (DriveNow), 95 did not use car sharing. Respondents were asked to use a custom-built app to track their trips and to specify the trip purpose and the mode of transport. The findings of this study confirm previous results obtained for station-based car sharing: free floating car sharing members have higher levels of education, higher income, fewer private cars, and more public transport subscriptions with respect to non-members. The study also highlights that car sharing members typically live in denser neighbourhoods, and are more intermodal and multimodal in their mobility behaviour. No statistically significant difference in trip purpose was detected between members and non-members: most trips are work-home trips ($57\%$), leisure ($19\%$) and shopping/errands ($13\%$). Giesel and Nobis~\cite{Giesel2016} perform a similar study for DriveNow and Flinkster users in Munich and Berlin, reporting substantially the same findings. 

Becker et al.~\cite{Becker2017} directly compare free floating and station-based car sharing members in the city of Basel, Switzerland. While the sociodemographic profile of car sharing is largely the same between station-based and free floating and substantially the same as that pictured in the previous literature, free floating car sharing members in Basel differentiate from their station-based counterpart in that they tend to use public transportation less. The authors remark that free floating car sharing may act as a complement to public transportation, filling the service gaps that their users might experience. The trip purposes of free floating car sharing members is quite diversified, but mostly involve visiting, shopping, and commuting, while station-based car sharing mostly covers leisure trips, goods transport, and shopping.

Wittwer and Hubrich~\cite{Wittwer2018} discuss the findings from a two-stage survey carried out in Hamburg, Germany, among Car2go members. The first stage of interviews took place in 2011, at the beginning of the Car2go service in the city, the second stage was run in 2016, when the service had been in place for a few years. From the sociodemographic standpoint, the 2011 and 2016 cohorts substantially share the same profile: largely man, 24-49 years old, high income, low car access, often with public transport season tickets. 2016 active users overwhelmingly rely on car sharing for leisure trips ($72\%$), but significant percentages also use it for shopping and errands ($50\%$) and for work/education trips ($42\%$). 

Based on the above overview, we can conclude that survey findings are consistent as far as the sociodemographic profiles of car sharing users are concerned, while contrasting results have been obtained regarding car sharing trip purpose and relationships with public transportation. In Section~\ref{sec:regression} we will discuss our findings in light of the above results.

\subsection{Knowledge mining from digital data}
\label{sec:relwork:digital_data}

The understandings and advancements brought about by the works described in Section~\ref{sec:survey} are invaluable, but the collection of survey data is expensive, time consuming, and does not scale. Typically, travel surveys cover a relatively small sample of all the trips of interest (because the the number of participants as well as the observation period are typically quite limited). Furthermore, it is a well-known problem that travel surveys often tend to underestimate the number of trips and to show a bias in the types of trips being reported~\cite{STOPHER2007}. For these reasons, in this work we depart from this approach and we exploit public, web-based, digital records, whose geotagged and time-stamped variety of data can be analysed with data mining techniques. These data can be collected for a possibly very long time with minimal effort, and can provide geographically diverse and almost continuous measurements of the systems under study. 

\color{black}

In the related literature, the works by Schm{\"o}ller et al.~\cite{Schmoller2015} and Willing et al.~\cite{Willing2017} are mostly focused on the external factors that may influence car sharing demand. In particular, Schm{\"o}ller et al.~\cite{Schmoller2015} highlight the role played by weather and demographics on the car sharing demand, while Willing et al.~\cite{Willing2017} tackle the problem of understanding if Points of Interest (PoI) in each city can be used as demand predictors. Differently from Willing et al.~\cite{Willing2017}, in this work we study the effects of PoIs taking into account collinearity of predictor variables and selection bias in p-value computation, resulting in a much smaller effect of PoIs on the car sharing demand. The same considerations apply for Schm{\"o}ller et al.~\cite{Schmoller2015}. Our work is also close to~\cite{Kortum2016}, which considers free-floating car sharing in multiple cities.  However, Kortum et al.~\cite{Kortum2016} focus on the growth rate of free floating car sharing rather than on the characterisation from the supply side point of view. Finally, in~\cite{boldrini2016characterising}, we have presented an analysis of station-based car sharing in a single city. The analysis in~\cite{boldrini2016characterising} is more oriented to issues related to the presence of stations (their capacity, how their behaviour can be mathematically modelled using queueing theory, etc.) and suffers from the lack of vehicle identifiers in the dataset. The technique used in~\cite{boldrini2016characterising} for detecting station usage is adapted here to the free floating case, but the analysis presented here is richer, because richer is the dataset extracted from the free floating car sharing operator.


\color{black}
Several works in the literature also focus on the problem of demand forecasting, which we tackle in Section~\ref{sec:forecasting}. This is typically done in conjunction with a proposal regarding vehicle relocation, which involves deciding how to proactively relocate shared vehicles in the operation area in order to meet the future demand. We can group forecasting proposals in three different classes, based on the approach they rely upon. 
There is a group of papers whose forecasting approach relies on techniques for \emph{time series} forecasting. Wang et al.~\cite{Wang2010}  leverage selective moving averages, Holt's model, Winter's model as well as Tabu Search heuristics for forecasting the demand in a car sharing service in Singapore. No prediction evaluation is carried out in the paper. M{\"u}ller and Bogenberger~\cite{Muller2015} focus on the city of Berlin and investigate how to predict future bookings using seasonal ARIMA model and exponential smoothing with Holt-Winters-Filter. 
The second class of forecasting methods are those coming from the \emph{machine learning} domain. Cheu et al.~\cite{Cheu2006}, for example, compare the forecasting performance of a neural network approach against that of Support Vector Regression, and find that the former provides better predictions. Neural networks have been later used also in \cite{Xu2007,Schulte2015,Alfian2017}. The third class of forecasting approaches relies on \emph{custom solutions} specific for the problem at hand. Boyaci et al.~\cite{boyaci2015optimization}, for example, compile origin-destination matrices by simply averaging the observations for different hours of the day, days of the week, and months of the year from real car sharing data. Weikl and Bogenberger~\cite{weikl2013relocation} devise a prediction algorithm based on finding clusters of behaviours for daily timeslots. In all the above works, the evaluation of forecasting performance is carried out considering only a single city.

 A preliminary analysis~\cite{boldrini2017sharing} of this dataset has been presented at KNOWMe'17, an ECML-PKDD workshop without copyrighted proceedings. In this extended version, we have added the sociodemographic study (Section~\ref{sec:regression}) and the demand forecasting analysis (Section~\ref{sec:forecasting}). In addition, we have added the analysis of the spatial autocorrelation of vehicle availability clusters (Section~\ref{sec:cell_usage}).


\color{black}

\subsection{Knowledge mining for other transportation systems}

From the methodology standpoint, this work is close to \cite{o2014mining,sarkar2015comparing,yang2016mobility,gast2015probabilistic}, in which bike-sharing, rather than car-sharing, systems have been analysed. Due to the different nature of the two systems, people use them differently, hence the results obtained for bike sharing systems cannot be applied directly to car sharing. However, similar methodologies can be exploited, e.g., to group stations based on how they are used by the customers.


This work is also orthogonal to the research efforts in the area of car pooling/ride sharing~\cite{trasarti2011mining,santi2014quantifying}. The idea of car pooling/ride sharing is that people may share a vehicle (be it a private or public vehicle, e.g., a taxi cab) to perform their trips.
Works in the area of car pooling typically focus on the amount of rides that can be shared, based on the historical or real-time trajectories of users, hence their focus is very different from that of this work. 

\subsection{Operation models for car sharing}

As one of the pillars of a smart transportation system, car sharing has recently been the subject of extensive research from the operational standpoint. The research activity on this area has focused both on short and long term strategic decisions. The latter involves problems like planning the station/parking infrastructure~\cite{correia2012optimization,boyaci2015optimization,biondi2016optimalplanning} or planning the recharging infrastructure. The former is focused on decisions such as when and how to redistribute shared vehicles~\cite{ijrr12_fluid,Kek2009,Febbraro2012,boldrini2017stackable} or when and how to recharge them~\cite{rottondi2014complexity,biondi2016optimalcharging}. 



To address the above problems, optimisation frameworks and operational decision tools for car sharing systems have been studied in the literature, but the proposed solutions have often been evaluated either on simulated scenarios~\cite{nourinejad2015vehicle,uesugi2007optimization} or using as input the demand (in terms of origin/destination matrix) obtained from surveys~\cite{Jorge2014,correia2012optimization}. On the contrary, the availability of a statistical characterisation of the general properties of real car-sharing systems, as well as a precise understanding of their emerging trends, is essential to both researchers and operators in order to design more effective decision support tools, and for the calibration and validation of simulations of car sharing systems. Thus, a data-driven analysis as that presented in this paper can be exploited to both  drive and evaluate solutions for the supply-side of car sharing.

\section{The dataset}
\label{sec:dataset}
The dataset comprises pickup and drop-off times of vehicles in 10 European cities for one of the major free-floating car sharing operator (Table~\ref{tab:dataset_summary}). For nine of these cities, data has been collected between May 17, 2015 and June 30, 2015. For \muenchen, data covers the period from March 11, 2016 to May 12, 2016. The data has been collected every $1$ minute using the available public API, which yields responses in the form of JSON files. Errors in the data collection process are due to technical problems on the booking website, in which cases corrupted entries have been discarded from the dataset. Each entry in the dataset describes the longitude-latitude position of available shared vehicles in the car sharing system, plus additional information. \textcolor{black}{Each entry in the dataset has the following structure:}
\begin{multline}
< \texttt{vin, date\_time, lon, lat, fuel, interior, exterior, engine} >,
\end{multline}
where \texttt{vin} is the unique identifier of a vehicle, \texttt{date\_time} contains the date and the time at which the available vehicle has been observed, \texttt{<lon, lat>} are the geographical coordinates, \texttt{<interior, exterior>} refer to the cleanliness of the vehicle, \texttt{engine} specifies where the vehicle is electric or not. Due to faulty GPS systems, the reported coordinates may be inaccurate. For this reason the dataset has been preprocessed and coordinates that are manifestly invalid (e.g., cars available in different countries) have been discarded. Data preprocessing and analysis has been carried out in R. 

\begin{table}[t]
\caption{Summary of dataset}
\centering
\begin{tabular}{l|r|r|l|l|l|l|cl}
City & \#Trips & \# Cars & Op. Area [$km^2$] & Cars/$km^2$ & Starts & Ends &  Duration\\
\hline \hline
\amsterdam & 49,901 & 349 & 59 & 5.880 & 2015-05-17& 2015-06-30  & 45 days\\
\hline
\berlin & 223,044 & 981 & 160 & 6.115 & 2015-05-17 & 2015-06-30 & 45 days\\
\hline
\firenze & 18,944 & 198 & 61 & 3.268 & 2015-05-17  & 2015-06-30  & 45 days\\
\hline
\copenhagen & 12,168 & 194 & 41 & 4.712 & 2015-05-17 & 2015-06-30  & 45 days\\
\hline
\milano & 156,080 & 686 & 120 & 5.737 & 2015-05-17 & 2015-06-30  & 45 days\\
\hline
\muenchen & 81,862 & 499 & 89 & 5.592 & 2016-03-11 & 2016-05-12  & 63 days\\
\hline
\roma & 99,515 & 584 & 90 & 6.480 & 2015-05-17  & 2015-06-30 & 45 days\\
\hline
\stockholm & 15,612 & 250 & 36 & 6.871 & 2015-05-17 & 2015-06-30  & 45 days\\
\hline
\torino & 25,091 & 299 & 53 & 5.646 & 2015-05-17  & 2015-06-30  & 45 days\\
\hline
\wien & 144,474 & 829 & 110 & 7.569 & 2015-05-17 & 2015-06-30  & 45 days\\
\end{tabular}
\label{tab:dataset_summary}
\end{table}%

Given the nature of our dataset, movements of cars have to be inferred from their unavailability during a certain time frame. Thus, when a car disappears from location A to later reappear at location B, we assume that the car has been picked up for a trip. We have no explicit way for distinguishing between regular customer trips and maintenance trips (e.g., cars that have been picked up by the car sharing operator for cleaning or repairing), as we simply observe a car disappearing from the map. 


In order to understand the main characteristics, in terms of mobility, of the ten cities in which the car sharing system under study is operating, we have extracted information (summarised in Table~\ref{tab:modal_share}) from the Eurostat's City Urban Audit database~\cite{urbanaudit}. 
Figure~\ref{fig:modal_split_pca} summarises the main transportation mode in each city as resulting from the Principal Component Analysis applied to the reported modal share. We can identify three main classes of cities: one in which motorised modes dominate, one in which public transport (PT) and walking are more important, and one in which people move prevalently by bike. 

\begin{table}[t]
\caption{Modal share in the 10 cities}
\centering
\begin{tabular}{l | r | r | r | r}
City & Bike (\%) & Walking (\%) & Public Transport (\%) & Motorized (\%) \\ 
  \hline \hline
\amsterdam{} & 22.00 & 4.00 & 30.00 & 44.00 \\ 
  \berlin{} & 9.90 & 7.20 & 41.00 & 38.50 \\ 
  \firenze{} & 4.04 & 7.68 & 19.10 & 66.82 \\ 
  \copenhagen{} & 36.12 & 5.84 & 29.23 & 26.40 \\ 
  \milano{} & 2.05 & 8.82 & 35.20 & 51.53 \\ 
  \muenchen{} & 10.10 & 6.60 & 45.20 & 37.40 \\ 
  \roma{} & 0.25 & 7.37 & 23.30 & 68.34 \\ 
  \stockholm{} & 7.00 & 15.00 & 43.00 & 33.00 \\ 
 \torino{} & 1.47 & 10.30 & 24.20 & 64.07 \\ 
  \wien{} & 7.00 & 27.00 & 39.00 & 26.97 \\ 
   \hline
\end{tabular}
\label{tab:modal_share}
\end{table}%
%
%



\color{black}
In terms of pricing structure, the policy implemented by the car sharing operator at the time the dataset was collected was quite simple: the rental price is a linear function of the rental time (the specific price per minutes varies across the ten cities in the range $[0.24,0.46]${\euro}cent/min). No surge pricing nor proximity-based pricing were implemented in the ten cities. Also, there were no incentives for customers to change their destination and to bring back cars to areas where cars where more in demand. A per-kilometre fee is applied only when the car is used for more than about 200km.
\color{black}

Finally, an interesting feature of this dataset is that it contains entries for two cities (\copenhagen{} and \stockholm{} in our analysis) for which the car sharing operator has now shut down service. \textcolor{black}{An index that is often used as a measure of car sharing success is the \emph{vehicle utilisation rate}, defined as the number of daily trips per vehicle. A higher value means that vehicles are used intensively in the city, hence the car sharing service is more profitable. Please note that long trips in which customers rent the shared vehicle for a long time are not the target of car sharing services but belong to the class of long-term rental. For this reason, the vehicle utilisation rate, with its ability to capture the short and frequent trips, is a direct measure of car sharing effectiveness. Figure~\ref{fig:vehicle_utilisation} shows the utilisation rate in the ten cities. It is clear how vehicles in some cities are much more utilised than in others, even 2-3 times more. It is also interesting to note that the vehicle utilisation rate is the lowest in the two cities (\copenhagen{} and \stockholm{}) where the service has been shut down months after we had collected this dataset. Remarkably, in \torino{} and \wien{} there is quite a lot of variability in the utilization rate. This is due to vehicles being injected or removed from the system during the data collection period.}

\section{Demand characterisation through sociodemographic indicators and urban diversity metrics}
\label{sec:regression}
In this section we focus on the demand, i.e., on the number of pickup requests observed in the different areas of a city, and we investigate how they are related to sociodemographic and urban fabric indicators. We discuss these indicators (which are the explanatory variables for our model) below, together with a brief description of the spatial unit of analysis considered in this section. 

\emph{Sociodemographic data:} Sociodemographic indicators characterise the population in the different areas of a city. For this analysis, we need a granularity finer than city level\footnote{Please note that EU countries are legally bound to provide census data to the Eurostat database at most at the level of NUTS 2 (regions). The actual database (\url{https://ec.europa.eu/eurostat/web/population-and-housing-census/census-data/database}) contains data up to NUTS 3 level (provinces) but this is not enough for our purposes. For this reason, we resorted to individually checking the countries' official institutes for statistics.}. We were able to find open census data with the desired spatial granularity for the cities of \firenze{}, \milano, \roma, and \torino{}. For their analysis, we focus on indicators related to the marital status, age group, educational attainment, employment status, and commuting habits.  The census data are obtained from the Italian National Institute for Statistics (ISTAT) and correspond to the 2011 Italian Census\footnote{\url{https://www.istat.it/it/archivio/104317}}. 


\emph{Urban fabric data:} The wealth of activities (cultural, commercial, recreational, etc.) taking place in a specific area is characterised using information about the Points of Interest (PoIs) collected from the location-based social network Foursquare\footnote{Through the Foursquare Places API it is possible to browse the venues in a certain geographic area. Since the standard API returns at most 50 venues per input area, each city is split into several browsing areas, whose size is properly dimensioned to ensure that all the available venues are acquired.}. When a user enters a new PoI, they are prompted to enter one of the first-level categories defined by the platform, which are Arts \& Entertainment, College \& University,  Event, Food, Nightlife Spot, Outdoors \& Recreation, Professional \& Other Places, Residence, Shop \& Service. We do not consider the category Event because events are generally limited in time, hence they typically do not overlap with our period of observation of the car sharing dynamics. Using this information, the urban fabric is characterised computing the number of PoIs (per category and overall) in each area. We also include a measure of the diversity of the urban fabric in an area by exploiting the concept of venues entropy introduced in~\cite{karamshuk2013geo-spotting}. The venue entropy of an area $a$ is obtained as:
\begin{equation}
e(a) = - \sum_{c \in \mathcal{C}} \frac{n_c(a)}{n(a)} \times \log \left( \frac{n_c(a)}{n(a)} \right),
\end{equation}
where $\mathcal{C}$ is the set of first-level Foursquare categories, $n_c(a)$ denotes the number of PoIs of category $c$ in the area, and $n(a)$ is the total number of PoIs in $a$. Intuitively, the entropy measures the uncertainty in predicting the category of a venue taken at random from the area, so the harder the prediction, the greater the diversity. 

\emph{Spatial unit of analysis:} We are constrained to use the smallest census area for which data are provided. In case of census areas that only partially cover the car sharing operation area, we consider the polygon resulting from their intersection and we rescale the sociodemographic indicators according to the percentage of overlapping. In order to have consistent estimates of the indicators inside each unit of analysis, we discard the census areas that overlap for less than $20\%$ with the operation area.

The pickups events, the PoIs and the entropy in the spatial units of analysis for the four cities are illustrated in Figure~\ref{fig:pois_map}.

\subsection{Explanatory analysis}
\label{sec:availability_long_term_exploratory}

\emph{Methods:} We investigate the relation between the total number of pickups ($y$) and the indicators discussed above (which we denote with $x_k$) using a multivariate linear regression model of the form:
\begin{equation}
y = \beta_0 + \beta_1 x_1 + \ldots + \beta_j x_j + \epsilon,
\end{equation}
where $\beta_0 \ldots \beta_j$ are the unknown parameters and $\epsilon$ is the error term. As expected for the kind of indicators that we are considering, multicollinearity is present in the data. In order to mitigate its negative effects, we use Lasso shrinkage~\cite{tibshirani1996regression} to estimate the coefficient of our linear regression\footnote{We use Lasso regression as implemented in the R package \texttt{glmnet}~\cite{friedman2010regularization}, using 10-fold cross validation for parameter estimation.}. Another advantage of Lasso is that it also perform subset selection, whereby a reduced set of predictors that have the greatest effect on the response $y$ is selected. In short, Lasso minimizes the residual sum of squares subject to the sum of the absolute value of the coefficients being smaller than a constant.

In standard linear regression, significance tests are used to test the statistical reliability of the rejection of the null hypothesis (i.e., that a coefficient $\beta_k$ is zero). Recently, a significance test for the Lasso regression has been proposed~\cite{tibshirani2014significance}, that factors in the selection bias related to the subset of selected predictors\footnote{As an example, running an Ordinary Least Square linear regression on the selected subset of predictor and calculating the p-values associated to the coefficients would yield a very optimistic estimate of the significance, due to the fact that the subset of predictors is not selected independently of the data.}. We will use this significance test to provide the p-values for the coefficients of our regression. 

As for the predictors that are skewed, we handle them applying a log transformation.

\emph{Results:} The cities for which we were able to obtain census data with the required granularity are \milano, \firenze, \roma, \torino. For \firenze, the census areas with a significant overlap with the operation area were to few to get statistically meaningful results, so we also discarded this city. The Lasso regression results for the remaining Italian cities are shown in Table~\ref{tab:regression_ita}.

In all Italian cities (Table~\ref{tab:regression_ita}), an high educational attainment in a certain area is significantly associated with an increased demand for car sharing in that area. Vice versa, a low educational attainment is associated with lower demand in both \roma{} and \torino{}. \textcolor{black}{This is largely in agreement with the findings from survey data discussed in Section~\ref{sec:survey}: car sharing users tend to be better educated than the general population, and this signal is strong enough to be detected by the correlation between the demand and the demographic composition of neighborhoods.}

Regularly commuting outside the reference municipality correlates negatively with car sharing demand (\milano{}, \roma{}). This is due to the fact that car sharing vehicles cannot be parked outside the operation area, hence they are not suitable for this type of commuting. They would be suitable, though, if paired with local public transport, using car sharing as a first/last-mile solution. This does not seem the case for \milano{} and \roma{} (the presence of transport facilities is not affecting the demand). It might be the case for \torino{}, as commuting outside the municipality is not considered a good predictor of the demand while a certain effect of the presence of transport facilities is detected. However, this effect is not statistically significant, hence this conclusion cannot be drawn from the data at hand. \textcolor{black}{Our analysis seem to confirm the complex relationship between free floating car sharing and public transport discussed in Section~\ref{sec:survey}: the synergy or friction between the two could be heavily dependent on local characteristics. An ad hoc analysis of this relationship would be an interesting follow-up work of the current investigation.} 

The marital status and age never correlate with the car sharing demand in the cities under study. \textcolor{black}{The latter result is in contrast with the findings based on survey data, where age always played a significant role in the profiling of car sharing users. One explanation could be that the age-related signal is weaker than the education-related one. Then, due to collinearity effects, the explanatory power of age is not considered sufficient by the Lasso. Another explanation is that age alone has never been explanatory, and its presence has always been due to its correlation with higher education attainments (in most OECD countries, young people are more educated than the elderly\footnote{\url{https://data.oecd.org/eduatt/population-with-tertiary-education.htm#indicator-chart}.}).}

In terms of urban fabric indicators, the presence of nightlife activities is associated with increased demand in all three Italian cities, while the presence of outdoor and recreational activities, as well as professional PoIs and residences, have a statistically significant effect in \milano{} only. \textcolor{black}{Thus, leisure seems to be the main motivation behind car sharing trips in the three cities. Work-related trips are significant only in \milano{}. When comparing these results with the survey-based findings summarised in Section~\ref{sec:survey}, no clear trend emerges. While leisure and work trips are a common finding, the signal associated with shopping activities goes completely undetected in these three cities. }


\begin{table}[tp]
\caption{Lasso regression for Italian cities. \textmd{Each cell contains the coefficient estimated by Lasso regression. Statistical significance is reported as $\textrm{(***)} =  p < 0.001$, $\textrm{(**)} = p < 0.01$, $\textrm{(*)} = p < 0.05$, $\textrm{(.)} = p < 0.1$.}}
\centering
\begin{tabular}{llll}
  \hline
 \multirow{2}{*}{Predictors} & \multicolumn{3}{c}{Coefficients} \\ \cline{2-4}
  & \milano & \roma & \torino \\ 
  \hline
Total population & - & - & - \\ 
  \# unmarried & - & - & - \\ 
  \# married & - & - & - \\ 
  \# separated & - & - & - \\ 
  \# widows & - & - & - \\ 
  \# divorced & - & - & - \\ 
  Age $<$ 5 & - & - & - \\ 
  Age 5-9 & - & - & - \\ 
  Age 10-14 & - & - & - \\ 
  Age 15-19 & - & - & - \\ 
  Age 20-24 & - & - & - \\ 
  Age 25-29 & - & - & - \\ 
  Age 30-34 & - & - & - \\ 
  Age 35-39 & - & - & - \\ 
  Age 40-44 & - & - & - \\ 
  Age 45-49 & - & - & - \\ 
  Age 50-54 & - & - & - \\ 
  Age 55-59 & - & - & - \\ 
  Age 60-64 & - & - & - \\ 
  Age 65-69 & - & - & - \\ 
  Age 70-74 & - & - & - \\ 
  Age $>$74 & - & - & - \\ 
  Age $>$6 & - & - & - \\ 
  \# with university degree & 0.22315 (***) & 0.13065 (***) & 0.22787 (***) \\ 
  \# with high school degree & - & - & - \\ 
  \# with middle school diploma & - & - & - \\ 
  \# with primary school diploma & - & -0.10589 (.) & -0.08575 (***) \\ 
  \# literate & - & - & - \\ 
  \# illiterate & - & - & - \\ 
  \# employed & - & - & - \\ 
  \# unemployed & - & - & - \\ \ 
  \# stay-at-home & - & - & - \\ 
  \# students & - & - & - \\ 
  \# other situations outside workforce & - & - & - \\ 
  \# commuting inside the municipality & - & - & - \\ 
  \# commuting outside the municipality & -0.1054700 (***) & -0.02632 (*) & - \\ 
  \# getting money & - & - & - \\ 
   \# PoIs & - & - & - \\ 
   PoIs entropy & - & 0.19928 (***) & - \\ 
   \# Arts \& Entertainment & - & - & - \\ 
   \# College \& University & - & - & 0.04625 \\ 
   \# Food & - & 0.24642 (*) & - \\ 
   \# Nightlife Spot & 0.2224 (**) & 0.13826 (.) & 0.28672 (***) \\ 
   \# Outdoors \& Recreation & 0.17047 (***) & 0.08842   & 0.04746  \\ 
   \# Professional \& Other Places & 0.17197 (*) & 0.26091 & 0.26261  \\ 
   \# Residence & 0.12921 (**) & - & - \\ 
   \# Shop \& Service & - & - & - \\ 
   \# Travel \& Transport & - & - & 0.02488  \\ 
   \hline
\end{tabular}
\label{tab:regression_ita}
\end{table}

\section{Demand forecasting}
\label{sec:forecasting}
In this section, we focus on the elements that influence the short-term behaviour of a car sharing system and we exploit them to forecast the demand. As we are interested in a finer spatial granularity (e.g. block level), we depart from the census areas used in the previous section. We thus need to identify a meaningful spatial unit to define car availability in a given area. In fact, differently from station-based car sharing, in free floating car sharing there is no natural ``aggregation" point for vehicles, which can be freely picked up and dropped off anywhere within the operation area. We can still perform a spatial analysis of car sharing usage by dividing the operation area into smaller cells and studying what is the behaviour, over time, in each of these cells. In this work we consider cells with side length 500m, which is the maximal walking distance typically accepted by car sharing users~\cite{weikl2013relocation,Herrmann2014}. 

\color{black}

\color{black}
%

Demand predictability is one of the crucial aspects for every transportation system. In car sharing, in particular, it is of utmost importance for vehicle redistribution, whose goal is in fact to proactively move vehicles in order to address the future demand. \textcolor{black}{In~\cite{boldrini2017sharing}, comparing the time series of empty cells over time against that of available vehicles, we have shown that there typically a lot of empty cells but at the same time there are also a lot of available vehicles. This situation hints at a strong concentration of vehicles in certain areas, vehicles that could be proactively moved to where the customers most need them. } Vehicle redistribution is typically performed periodically (e.g., every hour) and can be represented as a continuous cycling between three phases: i) the \emph{forecast} phase, when the expected pickups and drop-offs during the next relocation window are predicted; ii) the \emph{selection} phase, when the areas with vehicle surplus and vehicle deficit are identified and matched; and iii) the \emph{dispatching} phase, when the relocation workforce is assigned the previously defined relocation tasks~\cite{weikl2013relocation}.

\textcolor{black}{Our goal in this section is not to develop a new custom-built method for demand prediction in car sharing systems, but rather to compare state-of-the-art solutions that belong to different forecasting approaches (see Section~\ref{sec:relwork:digital_data} for the discussion on existing methods) in order to understand their performance in the ten cities under study. Indeed, while prior work on demand prediction has focused on individual cities, it is important to assess the robustness of the most representative methods to cope with the heterogeneity of travel behaviours and urban fabric. 
The target audience of this analysis are researchers working on designing optimised transport models for car sharing who might benefit from knowing what is the best, off-the-shelf, approach to prediction, so that they can focus their efforts on optimising the selection and dispatching phase discussed above. Similarly, third-parties developers will benefit from this type of analysis. For example, one could think to set up a service (similar in vein to OpenStreetCab~\cite{Noulas2015}, whose goal is to provide the best option price-wise between Uber and NYC taxies for a given trip) whereby the most \emph{reliable} car sharing service is recommended (e.g., one that guarantees that a car will be available in the evening when one drives back home). Third-parties apps will most likely have access only to the public data made available by the car sharing operators (similar to the data we are dealing with).}

\emph{Problem definition:} The goal of demand prediction is to establish the vehicle deficit/surplus at the cells. It can be described using the general formula we presented in~\cite{boldrini2017stackable}, which we discuss hereafter in a simplified version. If we denote with~$T$ the interval at which relocation is performed, every $T$ minutes the car sharing operator will compute, for each cell $i$, the expected balance $\hat{b}_i$ of vehicles at cell~$i$ for the next $T$ minutes, which can be described as follows:
\begin{equation}\label{eq:balance}
\hat{b}_i = v_i + \hat{drop}_i - \hat{pick}_i,
\end{equation}
where $v_i$ is the number of cars currently parked at station $i$, while $\hat{drop}_i$ and $\hat{pick}_i$ are, respectively, the forecast number of drop-offs and pickups in the next time interval. Please note that $v_i$ is a known quantity as it photographs the current situation at cell $i$. Instead,  $\hat{drop}_i$ and $\hat{pick}_i$ have to be estimated from what has happened in the past\footnote{Note also that, for the sake of clarify, in Equation~\ref{eq:balance} we are intentionally neglecting the contribution of relocated vehicles that have yet to arrive at the cell from the previous relocation interval. This does not affect the forecast results discussed in this section because this number would be known in advance anyway and, thus, would not be part of the prediction process.}. In the following, we show how statistical learning can help fill this gap and thus close the relocation cycle.

\color{black}

Let us focus on a tagged cell $i$ belonging to the set of all cells $\mathcal{C}$. We denote the set of days in our observation period with $\mathcal{D}$. Then, we divide each day $d \in \mathcal{D}$ in bins of length $T$ (i.e. we discretize time). The prediction problem at hand is a typical one: we have historical data (a set of $N$ observations) about pickup and drop-offs at cell~$i$ in each bin $t$ for each day in $\mathcal{D}$. We have to predict what will happen in each bin of the next days. In the following, we use the general term \emph{event} to denote either pickup or drop-off events.

\emph{Features:} For each cell~$i$, we extract the following features for prediction: 
\begin{itemize}
\item number of events $e_{(i,d,t)}$ observed in cell $i$ at time $t$ of day $d$
\item the time of the day (corresponding to bin $t$)
\item the day of the week (Sunday, Monday, etc.)
\item whether the day is a weekday or not
\item average number of events $\hat{e}_{(i,d,t)}$ observed at bin $t$ of day $d$ in the neighbouring cells (we consider 2-hop neighbours only).
\end{itemize}
%

\color{black}
\emph{Methods:} We use the first $80\%$ of the days in the dataset for training, and we predict the remaining $20\%$\footnote{Please note that standard k-fold cross validation cannot be performed with time series because time series data are not independent across time. The approach used in this paper is the same used in~\cite{yang2016mobility}.}. We set the time window~$T$ to 1 hour, implying that we want to forecast pickups and drop-offs happening in a one-hour time frame. We only consider cells that have more than 30 events during the observation period. Then, we run the prediction algorithms and we measure the prediction error in terms of Root Mean Squared Error (RMSE). 

\color{black}
We now define a set of relevant prediction techniques to be evaluated on the datasets at hand. It is important to point out that car sharing operators do not the disclose any detail on their approach to demand prediction. Thus, comparing against state-of-the-art industrial benchmarks is not an option. The first two solutions that we consider are simple baselines based on historical averages/medians. With regards to our discussion in Section~\ref{sec:relwork:digital_data}, the third one is representative of the class of time series prediction. Then, we pick two approaches for the machine learning category: neural networks (which have been already used in the literature for car sharing~\cite{Xu2007,Schulte2015,Alfian2017}), and Random Forest (which has been shown to be extremely effective when applied to bike sharing booking predictions~\cite{yang2016mobility,sarkar2015comparing}). Finally, we test a technique in the custom forecasting category, specifically the one proposed in~\cite{weikl2013relocation}. In the following we provide a description of each technique.

\color{black}

\textbf{Prediction based on Historical Average (HA)}: this prediction function returns the average number of events observed in the same time window across different days. \textcolor{black}{In other words, the predicted number of events $\hat{y_t}$ at a certain time $t$ in the future is obtained as $\hat{y_t} = \frac{1}{|\mathcal{D}|}\sum_{d\in \mathcal{D}}e_{(i,d,t)}$. As car sharing typically exhibits marked differences between weekdays and weekends~\cite{boldrini2016characterising}, we also test a version of the algorithm (denoted as HA+)  that distinguishes between working days and weekends.} A similar function has also been used as benchmark in the related literature on bikesharing forecasting~\cite{yang2016mobility,gast2015probabilistic}.

\textbf{Prediction based on Historical Median (HM)}: the prediction function returns the median number of events observed in the same time window across different days\textcolor{black}{, i.e., $\hat{y_t} = \textrm{median}_{d\in \mathcal{T}}(e_{(i,d,t)})$}. This function is expected to perform well in cases where the distribution of pickups/drop-offs is highly skewed. As for the previous algorithm, we also test a version (denoted as HM+)  that distinguishes between working days and weekends.

\textbf{ARIMA}: the Autoregressive Integrated Moving Average technique is a popular time series forecasting method. It is a generalisation of the ARMA model used in~\cite{gast2015probabilistic,Muller2015}. Typically, ARIMA models are denoted with $\textrm{ARIMA}(p,d,q)$, where $p$ is the order (number of time lags) of the AR component, $d$ is the degree of differencing, and $q$ is the order of the MA component. \textcolor{black}{Here we use the seasonal version of the above ARIMA model, estimating the parameters for both the non-seasonal and the seasonal component (this allows us to detect cyclic behaviour, if it exists). \textcolor{black}{We remind that in a seasonal ARIMA model, seasonal AR and MA terms predict the target variable using data values and errors at times with lags that are multiples of $S$ (the span of the seasonality).} For each cell the best configuration of the ARIMA parameters is selected according to their Corrected Akaike Information Criterion (AICc) value, using the \texttt{auto.arima} function of R's \texttt{forecast} package. The search range for the parameters is the default one in the \texttt{auto.arima} function. Being this a time series method, only the temporal information of each observation and the actual observed values are fed to the model.}

\textbf{Random Forest (RF)}: tree-based learning method that aggregates the prediction results of several decision trees obtained by randomly selecting, each time, only a subset $m$ of the original $p$ features (those described in the features section above). \textcolor{black}{In order to select the most appropriate $m$, we used 5-fold cross validation and we vary\footnote{Note that the set of initial features ($p=3$) is expanded after applying OneHot encoding. For example, the categorical day of the week is split into 6 binary features.} $m$ in $\{2,4,5\}$. We use the implementation in the R package \texttt{randomForest}, together with the \texttt{caret} package for training and prediction.}
Random Forest was found to be the best prediction function for bikesharing in~\cite{yang2016mobility}.

\color{black}
\textbf{Neural Network (NN):} relying on the same settings as in~\cite{Cheu2006}, we use a single layer perceptron with as many neurons in the input layer as the features described above, one hidden layer (searching for the best number of neurons between 1 and 30), single output neuron,  backpropagation, hyperbolic tangent activation function, linear output function. Categorical features have been represented using dummy variables. Then, input and output data were scaled to the range $[-1,1]$, which is the sensitive range of the hyperbolic tangent activation function. We rely on the implementation in R package \texttt{RSNNS}, together with the \texttt{caret} package for training and prediction. Parameters selection is again performed using 5-fold cross validation.

\textbf{Algorithm in Weikl and Bogenberger~\cite{weikl2013relocation} (WEIKL):} one of the very few custom proposals in the literature on car sharing, the rationale of this algorithm is to represent each timeslot of each day through a vector, whose components are the number of events at each cell during the timeslot. Let us focus on a tagged timeslot $t$. These vectors describing the spatial demand for timeslot $t$ across each day make up a matrix of size $|\mathcal{C}| \times | \mathcal{D_{T}}|$ (where $\mathcal{C}$ denotes the set of cells and $\mathcal{D_{T}}$ denotes the set of days in the training set). The $|\mathcal{C}|$-dimensional representation of each day is then simplified using Principal Component Analysis, and only the first two principal components are retained. This two-dimensional description of the days is then clusterised using $k$-means, in order to group together days featuring the same demand behaviour. In the original paper, how the optimal number of groups is obtained is not specified, so we decided to rely on the gap statistic~\cite{Tibshirani2001}, a state-of-the-art solution that is able to handle also the single-group case (i.e., to detect when the optimal choice is to not split in groups). Once this has been done for all timeslots, a so-called from-to matrix is built, computing the probability that days in a certain group $g_i$ in timeslot $t$ would be in group $j$ in timeslot $t+1$. Using this from-to matrix, it is possible to compute the demand variation from a timeslot to another for each group. This concludes the training phase of the algorithm. In the prediction phase, the demand in timeslot $t-1$ is mapped into one of the groups computed in the training phase (by closest centroid matching). Then, the number of forecasted events for timeslot $t$ is obtained from the computed expected demand variation for the group. Please note that in~\cite{weikl2013relocation}, each day was divided in timeslots of non-uniform size. For fairness with the other prediction algorithms, we use timeslots of fixed size $T$. We have implemented this method in R.
\color{black}


\emph{Results:} The results are shown in Figures~\ref{fig:mean_rmse_60_pickup} and~\ref{fig:mean_rmse_60_dropoff}, for pickups and drop-offs respectively. For most cities and for all algorithms, the error is small, with forecasts off, on average, by less than one drop-off/pickup for the vast majority of cells. However, there are a few cells for which the prediction error is high. After an in-depth analysis of the nature of these cells, we discovered that they are typically in very busy areas (e.g. near the airport), where both the high volume of traffic and the bustier nature of arrivals and departures may explain this variability. Also, the RMSE for pickups tends to be slightly higher than for drop-offs. 
\color{black}
In terms of which prediction algorithm works best, Figure~\ref{fig:best_predictors_60} shows that Random Forest provides the most accurate predictions for the vast majority of cells. The WEIKL algorithm is the second best, but its performance is very close to that of the NN approach and, surprisingly, to the simple Historical Average. HA+ and HM+, the versions of HA and HM algorithms that take into account the difference between weekdays and weekends, do not outperform in general their simpler counterparts. ARIMA, used also in~\cite{Muller2015} for forecasting car sharing demand, provides consistently the worst predictions. 
\color{black}

In Figure~\ref{fig:taggedCell_predicted_ts}, in order to showcase the main strengths and weaknesses of the prediction techniques used, we focus on a tagged cell (specifically, on one for which the error is generally large) and we plot the time series of \textcolor{black}{the predicted drop-offs (black curve) against the observed drop-offs (blue and red, in order to distinguish between weekdays and weekends). For the sake of readability, we consider one strategy per class of prediction approach: HA for the simple baselines, ARIMA for the time series forecasting class, RF for the machine learning approaches, and WEIKL for the custom solutions.}
The ARIMA model tends to replicate the same daily patterns across all days in the test set, since the ARIMA model is not able to capture multiple seasonalities, which are instead present in the data. By using predictive models that explicitly handle these multiple seasons (such as~\cite{de2011forecasting}), the quality of prediction could be significantly improved. A similar problem seems to hold for HA: it tends to replicate a ``model day'', which is always the same. Instead, the predictions provided by the Random Forest algorithm are the most flexible ones, as they seem to adapt individually to each day. However, despite this flexibility, there seems to be an inherent variability in certain cells in the datasets (Figures~\ref{fig:mean_rmse_60_pickup}-\ref{fig:mean_rmse_60_dropoff}) that makes prediction difficult. \textcolor{black}{The tagged cell considered here is also useful to illustrate the weakness of the WEIKL solution. Since it groups together many cells to extract a typical behaviour of the system in a given timeslot, the cells with a small number of events (which are many) tend to dominate over the more active ones (like the tagged cell considered here). Thus, in these cases, the predictions are significantly off with respect to the actual behaviour of the cell.}

\color{black}
\section{Spatiotemporal usage patterns}
\label{sec:cell_usage}

It is expected that cells in a car sharing system are used differently by the users, but how many different usages can be identified? In order to answer this question, in the following we carry out a classification of cells based on their usage pattern. To this aim, we focus on the time series of vehicle availability in each cell and we measure how close this time series is with what we observe in other cells. We measure the time series distance using the Dynamic Time Warping (DTW) technique~\cite{esling2012time} (with Sakoe-Chiba band), then we cluster cells based on their DTW-distance using Partition Around Medoids (PAM) clustering. For each city, the optimal number of clusters is obtained using the silhouette method. In order to be able to compare our time series, we discretise time into bins with a duration of 10 minutes. For each cell, we extract one availability value per bin by averaging the availability in the bin in different days. In addition, in order to detect variation above and below the average behaviour, we normalise the measured availability using the average availability at the cell.

The results are shown in Figure~\ref{fig:clusters_ts}. \textcolor{black}{The optimal number of clusters is 2 in \amsterdam{}, \firenze{}, and \copenhagen{}, 3 in \berlin{}, \milano, \roma{}, \stockholm, \torino{}, and 4 in \muenchen{} and \wien{}}. However, the fourth cluster, when present, is a very special cluster, composed of just a single cell. This single cell is a very special one in the city ecosystem, and in both cities where the fourth cluster is present, this cluster comprises the airport zone. If we plot the availability time series within each cluster (Figure~\ref{fig:clusters_ts}, obtained by computing the average availability in the cells belonging to the cluster), it is striking to see that the clusters highlight very characteristic cell usage. Some cells have above average availability at night and below average availability during the day. Other cells have exactly the opposite behaviour. Finally, there is a group of cells with an intermediate behaviour, where apparently no significant difference in usage is detected over the whole day. It is easy to map this behaviour into the ``nature'' of the area covered by the cell: people leave residential areas in the morning and come back in the evening, while the opposite is true for commercial/business areas. Similar classes were identified in~\cite{boldrini2016characterising} for station-based car sharing, and in~\cite{sarkar2015comparing} for bike sharing.
Figure~\ref{fig:clusters_ts} also highlights the outlier behaviour of the airport zone (which constitutes the fourth cluster, when available). Airports in \muenchen{} and \wien{} see a huge variation in availability; however, the behaviour of their time series is simply a scaled version of the commercial/business pattern discussed before. \textcolor{black}{Due to the magnitude of the airport clusters' time series, the behaviour of the other clusters of \muenchen{} and \wien{} is barely visible in the plot. If we zoomed in, we would see the typical patterns that can be seen more clearly in cities with no airport within the operation area.}

Based on the above discussion, we can associate each cluster with the trend in its corresponding availability time series. Thus, we identify four main behaviours: cells with mid-of-day availability peak, cells with night peak, cells with no significant peak, and cells whose availability variations are much higher than in other cells. We use the labels \emph{day}, \emph{night}, \emph{neutral}, and \emph{high-intensity} to refer to these four classes. In the following, we investigate to which extent the behaviour of cells is spatially autocorrelated. To this aim, since cell labels are categorical, we use the Join Count statistics~\cite{cliff1981spatial}. With this approach, for each cell $n$, we count how many of its neighbouring cells belong to $n$'s class and we compare this result with what would be obtained if classes were distributed uniformly at random across cells. Since the high-intensity class comprises at most one cell per city, we discard it from the analysis. The results for all cities are shown in Table~\ref{tab:joincount}. Cells exhibiting an availability peak at night are spatially autocorrelated in all ten cities. Cells with a mid-of-day peak are spatially correlated in all cities except for \firenze{} and \copenhagen. Out of the seven cities featuring neutral cells, the spatial autocorrelation is significant for only three of them. We can conclude that, in general, the availability of vehicles in cells tends to be spatially autocorrelated, hence neighboring cells tend to have shortage/abundance of vehicles at the same time. \textcolor{black}{This further motivates the use of vehicle availability information in neighbouring cells for demand forecasting (RF and NN in Section~\ref{sec:forecasting} indeed rely on this information and their performance is quite good, with RF being the most performing prediction algorithm overall).}

%
%

\begin{table}[ht]
\caption{Join Count Statistics. \textmd{Column \emph{Count} contains the total number of matches; column \emph{Exp. (rand)} contains the number of matches expected under random;  \emph{Test Stat.} contains the test statistics;  $^{(\dagger)}$ denotes p-values greater than 0.05 (not statistically significant). Neighbour cells are obtained using the Queen criterion (i.e., assuming moves like the Queen in a chess game); spatial weights are binary.}}
\centering
\begin{tabular}{llrrrl}
  \hline
City & Cluster Type & Count & Exp. (rand) & Test Stat. & p-value \\ 
  \hline
\multirow{2}{*}{\amsterdam{}} & day & 753 & 726.56 & 2.22 & 1.33e-02 \\ 
 & night & 132 & 53.13 & 12.40 & 1.25e-35 \\ \cline{2-6}
\multirow{ 3}{*}{\berlin{}} & neutral & 1623 & 1571.39 & 2.44 & 7.42e-03 \\ 
 & night & 501 & 295.93 & 14.81 & 6.38e-50 \\ 
 & day & 254 & 65.88 & 25.53 & 4.31e-144 \\ \cline{2-6}
\multirow{ 2}{*}{\firenze{}} & day & 400 & 403.76 & -0.33 & 6.29e-01 $^{(\dagger)}$\\ 
  & night & 109 & 86.40 & 2.95 & 1.60e-03 \\ \cline{2-6}
\multirow{ 2}{*}{\copenhagen{}} & day & 385 & 385.50 & -0.05 & 5.22e-01 $^{(\dagger)}$\\ 
  & night &  86 & 59.11 & 4.37 & 6.18e-06 \\ \cline{2-6}
\multirow{ 3}{*}{\milano{}} & neutral & 837 & 793.31 & 2.82 & 2.40e-03 \\ 
  & night & 349 & 174.15 & 16.54 & 9.25e-62 \\ 
  & day &  89 & 10.18 & 26.01 & 1.92e-149 \\ \cline{2-6}
\multirow{ 3}{*}{\muenchen{}} & neutral & 613 & 588.51 & 1.64 & 5.10e-02 $^{(\dagger)}$\\ 
  & night & 221 & 98.37 & 14.37 & 3.95e-47 \\ 
  & day &  82 & 26.10 & 11.79 & 2.13e-32 \\ \cline{2-6}
\multirow{ 3}{*}{\roma{}} & neutral & 623 & 609.62 & 0.90 & 1.84e-01 $^{(\dagger)}$\\ 
  & night & 163 & 88.58 & 9.24 & 1.26e-20 \\ 
  & day & 265 & 75.74 & 25.09 & 3.35e-139 \\ \cline{2-6}
\multirow{ 3}{*}{\stockholm{}} & neutral & 192 & 198.33 & -0.69 & 7.56e-01 $^{(\dagger)}$\\ 
  & night &  66 & 50.95 & 2.53 & 5.72e-03 \\ 
  & day &  22 & 9.13 & 4.60 & 2.10e-06 \\ \cline{2-6}
\multirow{ 3}{*}{\torino{}} & neutral & 422 & 408.32 & 1.26 & 1.04e-01 $^{(\dagger)}$\\ 
  & night & 152 & 100.44 & 6.61 & 1.96e-11 \\ 
  & day &  41 & 4.01 & 19.43 & 2.20e-84 \\ \cline{2-6}
\multirow{ 3}{*}{\wien{}} & night & 372 & 231.90 & 11.83 & 1.29e-32 \\ 
  & neutral & 657 & 599.45 & 3.75 & 8.80e-05 \\ 
  & day &  85 & 18.59 & 16.45 & 3.90e-61 \\ 
   \hline
\end{tabular}
\label{tab:joincount}
\end{table}

\section{Locating cleaning and maintenance areas}
\label{sec:service_areas}
A critical operational aspect for car sharing is how to perform cleaning and maintenance. \textcolor{black}{When not done properly, it may even be a critical factor of the service shutdown, as in the case of Parisian car sharing Autolib}\footnote{\url{https://www.thelocal.fr/20180619/wheels-set-to-come-off-paris-autolib-electric-cars}}. In order to perform cleaning and maintenance, the car sharing workforce is typically remotely dispatched to collect vehicles that are in need of either. However, moving workers around is expensive, and more efficient solutions could be found based on the vehicle usage in the city. As a case study, in the following we discuss how to identify potential service areas within the operation area. A potential service area is a location vehicle pass by with very high probability. A workshop could be  deployed in this area, and this would make cleaning and maintenance operations much more efficient. 

We can use our dataset to understand if these potential service areas exist or not in the cities covered by the car sharing service under study. To this aim, we define a reference window $W$, corresponding to the accepted tolerance for taking out a vehicle for maintenance. Based on data from active car sharing operators, we assume that reasonable values for $W$ are between 15 and 30 days. Then, for each cell, we count the number of distinct vehicles seen by the cells during $W$. Figures~\ref{fig:service_areas_w30} and~\ref{fig:service_areas_w15} show the results for the top three cells in each cities, i.e., the three cells that see the highest number of distinct vehicles during two different time windows ($W=30$ and $W=15$ days, respectively).  \textcolor{black}{Assuming that a (somewhat generous) threshold of $50\%$ vehicles would be acceptable for the car sharing operator to justify the opening of a workshop in the area, all cities with the exception of \firenze{} would accomodate three workshops satisfying this requirement when $W=30$. The scenario $W=15$ is by far more challenging: six cities would be able to open at least one workshop, but only one city could open two and three. The top ranking cell for cities whose operation area covers the airport is always the cell that includes the airport, which thus becomes a strategic asset in car sharing operations, in addition to being a huge generator of car sharing traffic. }

\color{black}
\section{Conclusions}


In this work, we have collected web-based data about free floating car sharing in 10 European cities, cities that are heterogeneous both in terms of car sharing success and mode split. We have studied how the car sharing demand relates to sociodemographic and urban indicators, showing that the car sharing demand is positively correlated with high educational attainment and nightlife activities, while being negative correlated with the percentage of people commuting outside the municipality. These findings both confirm and extend the results in the related literature obtained from survey data. Then, focusing on the predictability of future car sharing requests, we have shown that they can be forecasted quite accurately using state-of-the-art prediction algorithms, and we have highlighted the very good performance of Random Forest as predictor. Finally, we have proposed a strategy for selecting the area in which maintenance facilities should be deployed, and we have shown how the airport zone can become a strategic asset for car sharing operators, due to the fact that the high volume of traffic generated by the area makes it extremely convenient to deploy cleaning and maintenance facilities there.

\begin{backmatter}

\section*{List of abbreviations}

PT, Public Transport; RMSE, Root Mean Squared Error; PAM, Partition Around Medoids; AICc, Corrected Akaike Information Criterion;

\section*{Availability of data and materials}

A sample of the dataset used in the paper will be available at publication time at the following link \url{https://github.com/chibold/carsharing}.

\section*{Competing interests}
  The authors declare that they have no competing interests.

\section*{Funding}
This work was funded by the ESPRIT, REPLICATE and SoBigData projects. The ESPRIT project has received funding from the \emph{European Union's Horizon 2020 research and innovation programme} under grant agreement No 653395. The REPLICATE project has received funding from the \emph{European Union's Horizon 2020 research and innovation programme} under grant agreement No 691735. The SoBigData project has received funding from the \emph{European Union's Horizon 2020 research and innovation programme} under grant agreement No 654024.

\section*{Author's contributions}
Designed the study: CB RB. Analyzed the data: CB HL. Wrote the paper: CB RB. All authors read and approved the final manuscript.

\section*{Acknowledgements}
Not applicable.

\section*{Authors' information}
CB and RB are permanent researchers at IIT-CNR. HL was a postdoctoral researcher at IIT-CNR at the time the study was carried out.


\bibliographystyle{bmc-mathphys} 
\bibliography{epj_ds18}      




\section*{Figures}

\begin{figure}[!h]
\centering
\iffigures\includegraphics[scale=0.3]{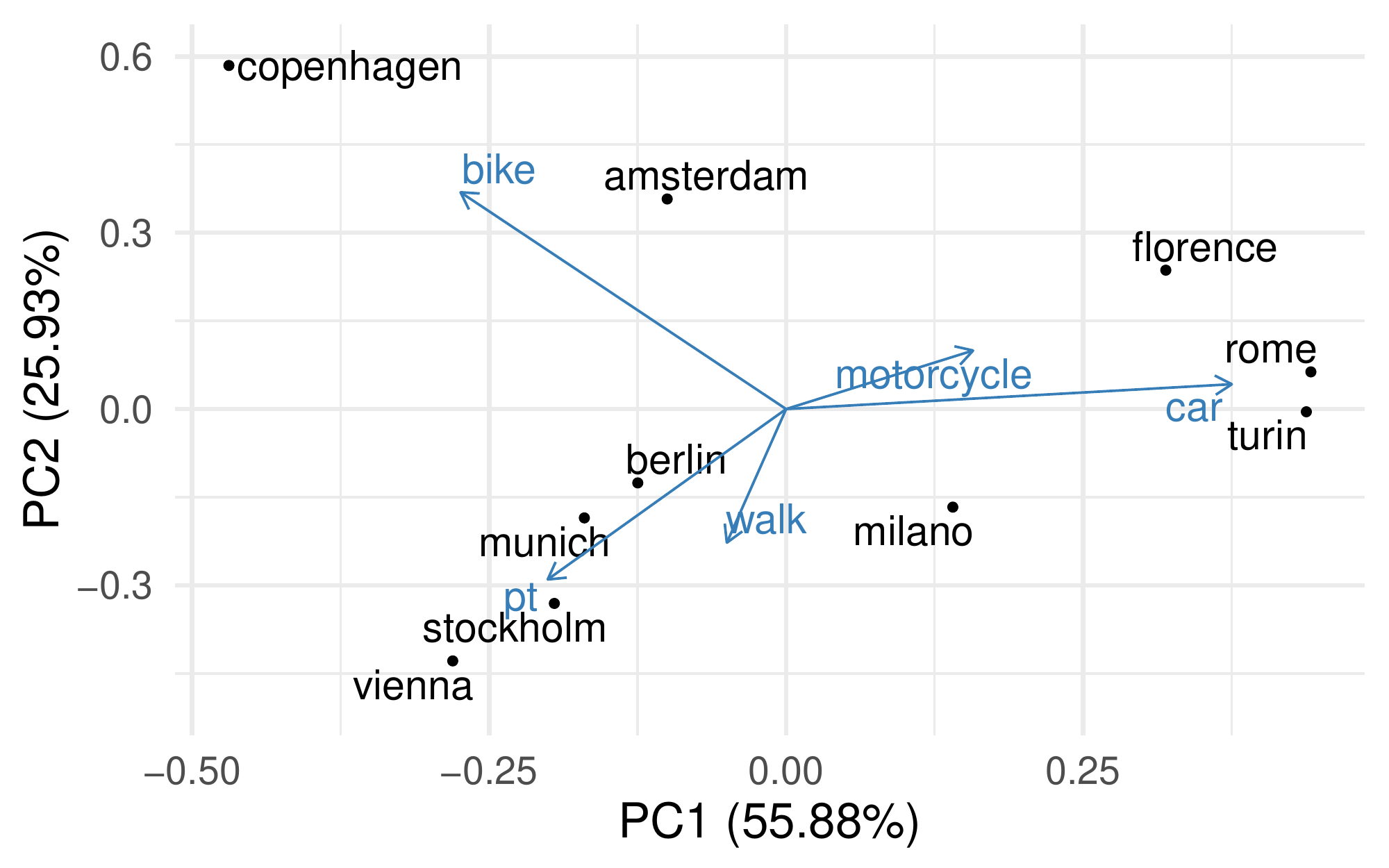}\fi
\caption{\csentence{Modal split: Principal Component Analysis.}}
\label{fig:modal_split_pca}
\end{figure}

\begin{figure}[!h]
\centering
\iffigures\includegraphics[scale=0.4]{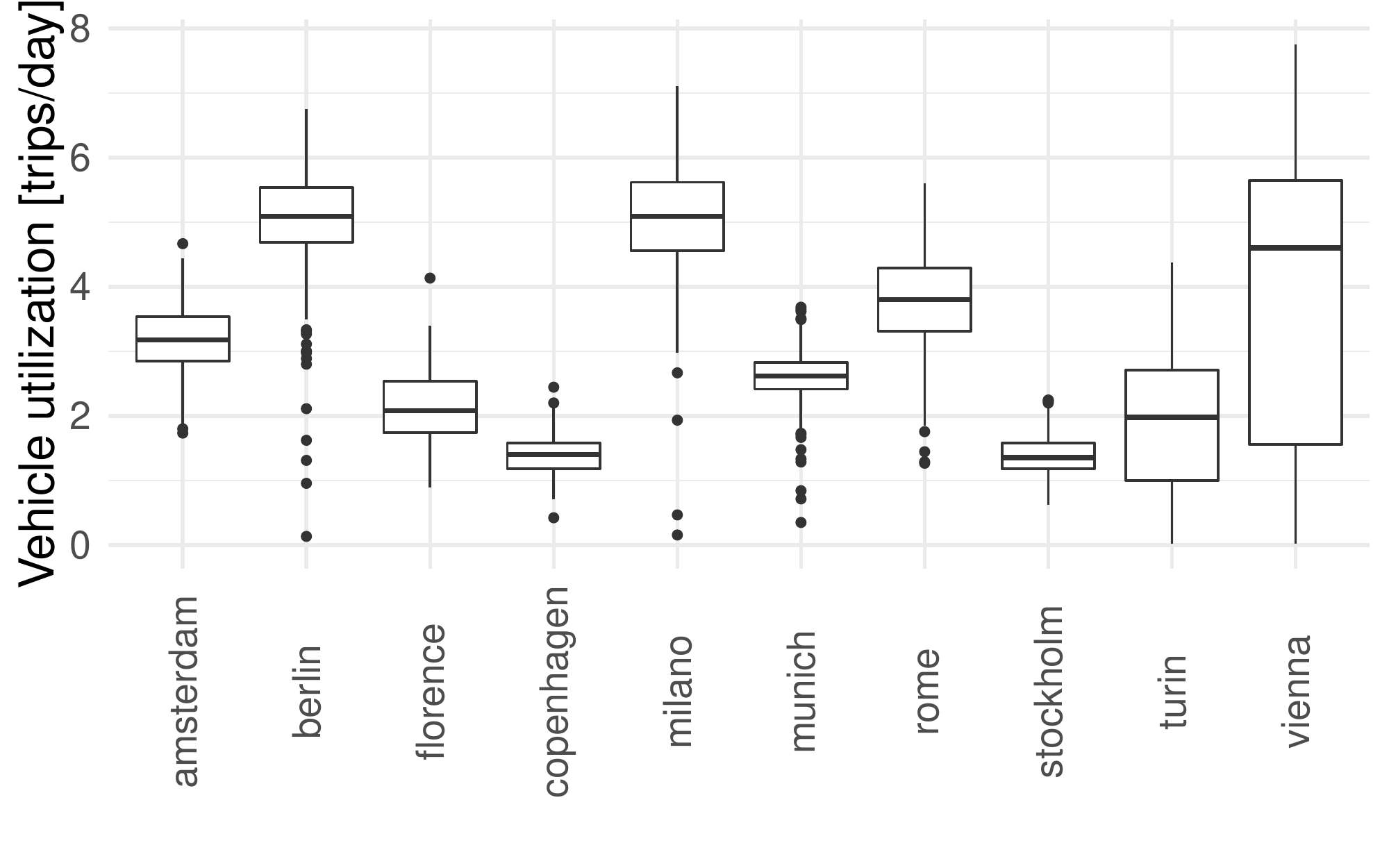}\fi
\caption{\csentence{Utilisation rate for vehicles in the shared fleet. \textmd{The boxplots compactly display the distribution of the utilisation rate across the cells of the ten cities.}}}
\label{fig:vehicle_utilisation}
\end{figure}

\begin{figure}[!h]
\centering
\iffigures\includegraphics[scale=0.3]{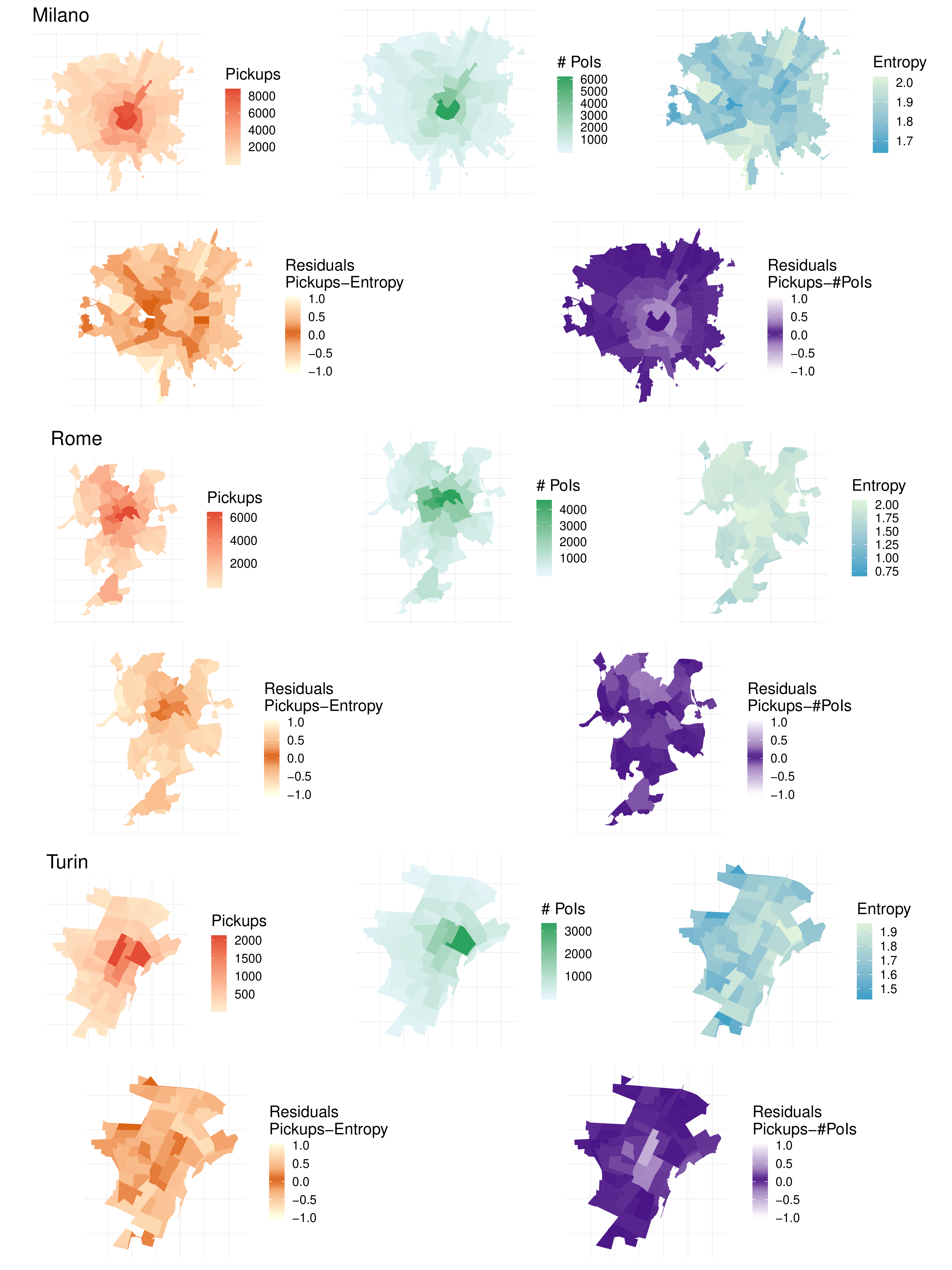}\fi
\caption{\csentence{Map of number of pickups, number of PoIs, entropy, residuals of rescaled pickups and entropy, residuals of rescaled pickups and PoIs in \milano{}, \roma{}, \torino{}.} \textmd{The residuals are computed after rescaling the variables in range $[0,1]$, then subtracting. Intuitively, they correspond to the deviation from perfect correlation.}}
\label{fig:pois_map}
\end{figure}

\begin{figure}[!h]
\centering
\iffigures\includegraphics[scale=0.5]{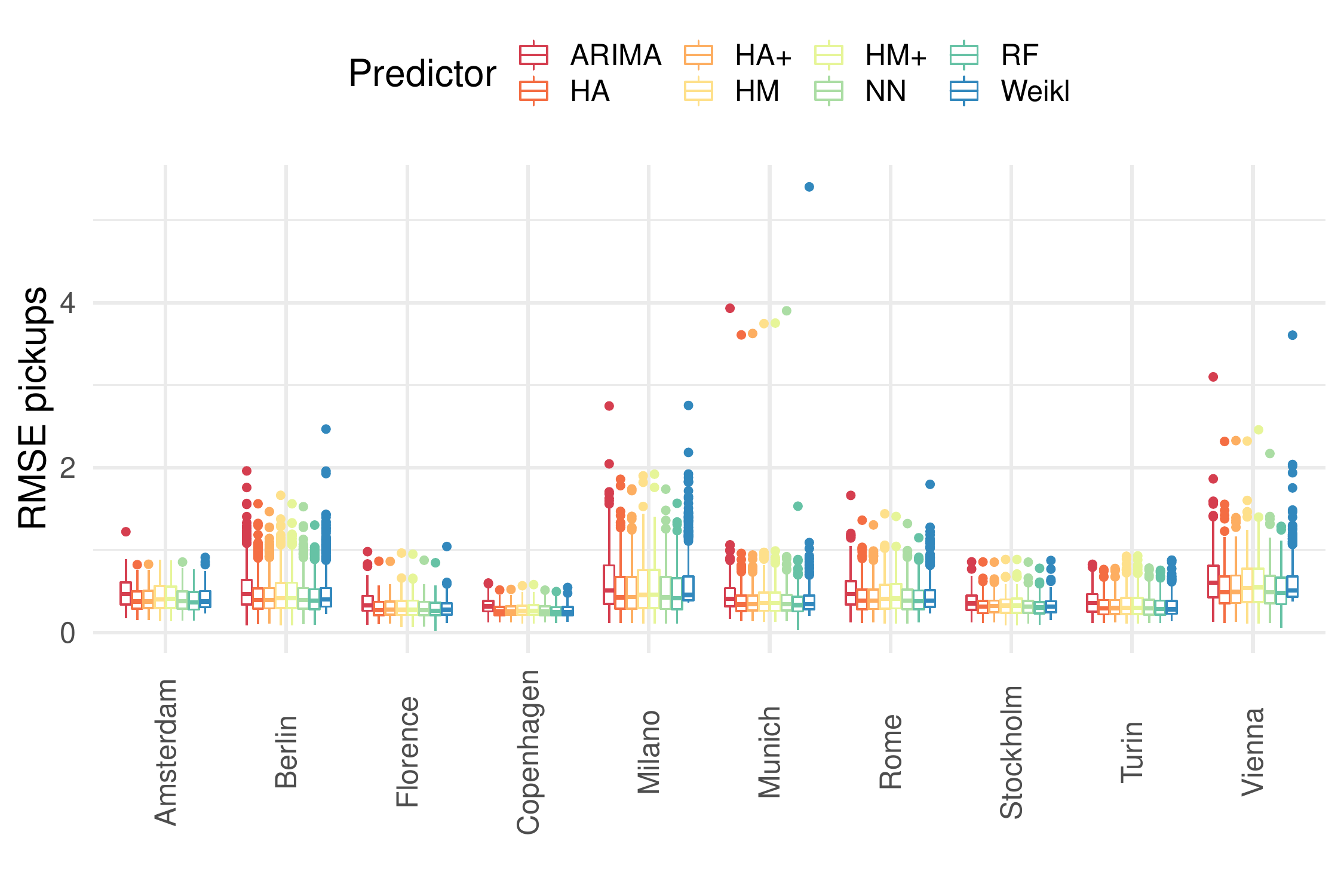}\fi
\caption{\csentence{Box plot of RMSE for pickups in the 10 cities (1 hour time window).}}
\label{fig:mean_rmse_60_pickup}
\end{figure}

\begin{figure}[!h]
\centering
\iffigures\includegraphics[scale=0.5]{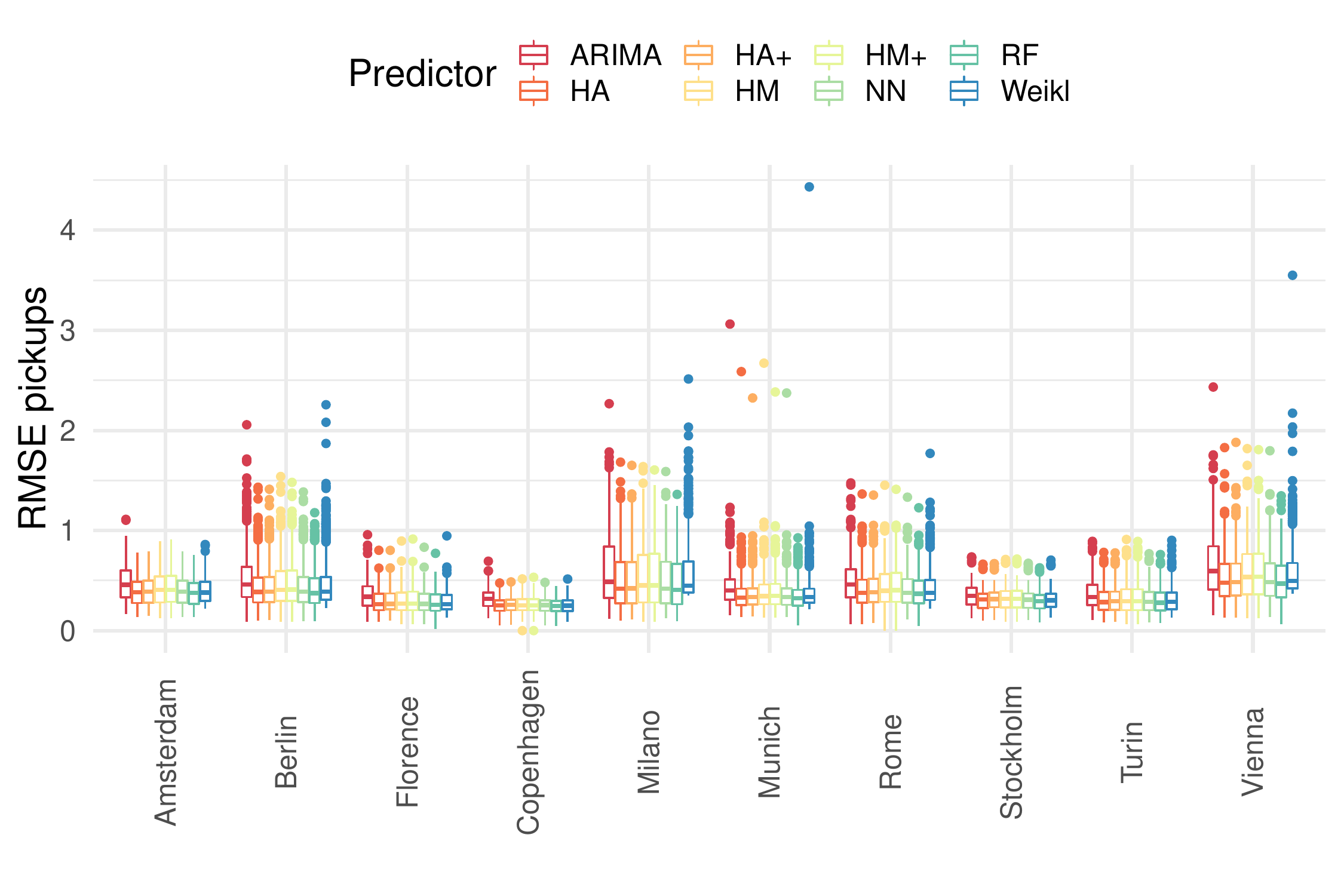}\fi
\caption{\csentence{Box plot of RMSE for drop-offs in the 10 cities (1 hour time window).}}
\label{fig:mean_rmse_60_dropoff}
\end{figure}

\begin{figure}[!h]
\centering
\iffigures\includegraphics[scale=0.6]{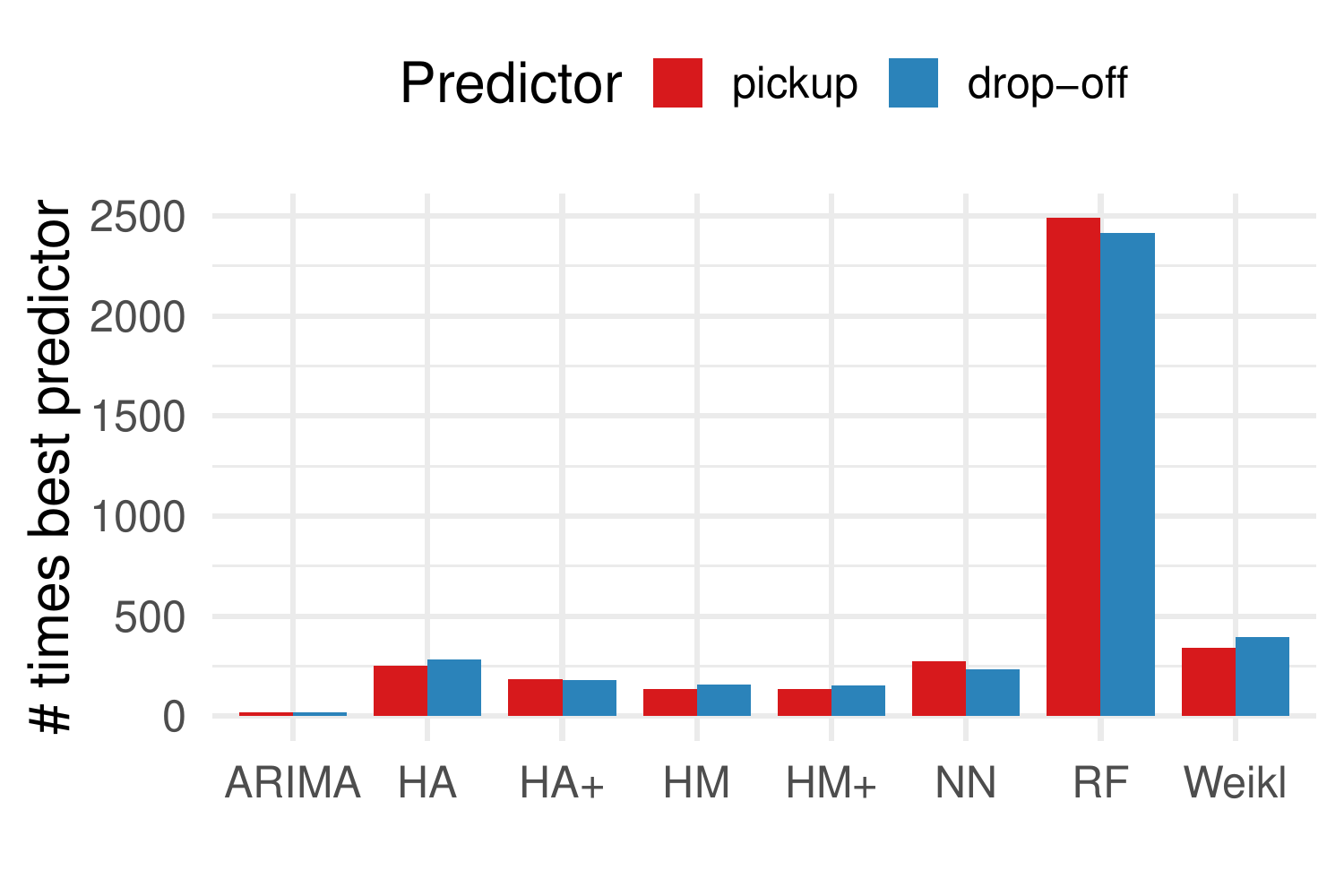}\fi
\caption{\csentence{Best predictors (1 hour time window).} The y-axis shows the number of times, over all predictions tasks, that each predictor has outperformed the others.}
\label{fig:best_predictors_60}
\end{figure}

\begin{figure}[!h]
\centering
\iffigures\includegraphics[scale=0.45]{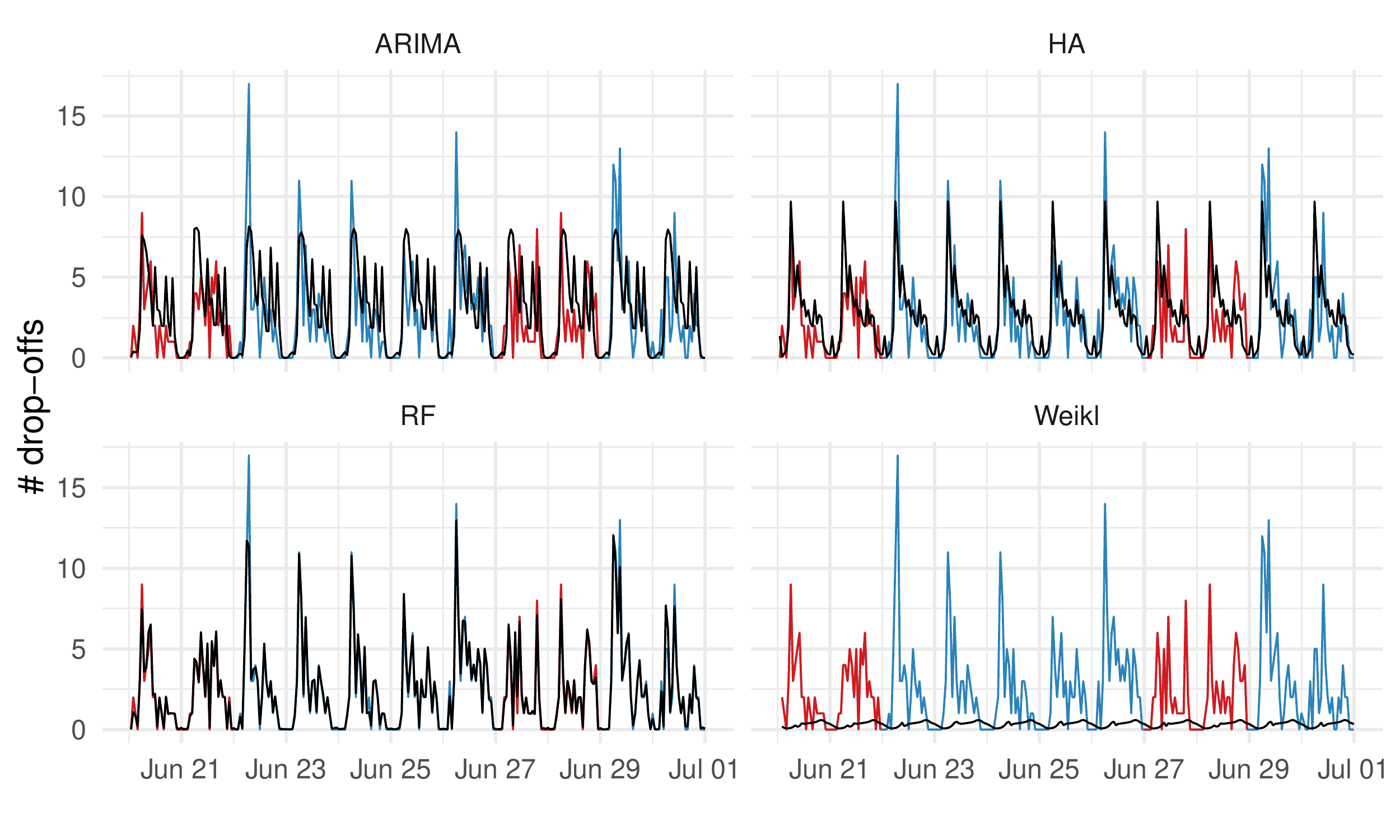}\fi
\caption{\csentence{Time series of predicted vs observed drop-offs for a tagged cell.} RMSE: 2.44 (ARIMA), 1.83 (HA),  1.72 (RF), 3.56 (WEIKL). Black is used for the predicted time series, while blue and red are for weekdays and weekends, respectively, of the test dataset.}
\label{fig:taggedCell_predicted_ts}
\end{figure}

\begin{figure}[!h]
\centering
\iffigures\includegraphics[scale=0.45]{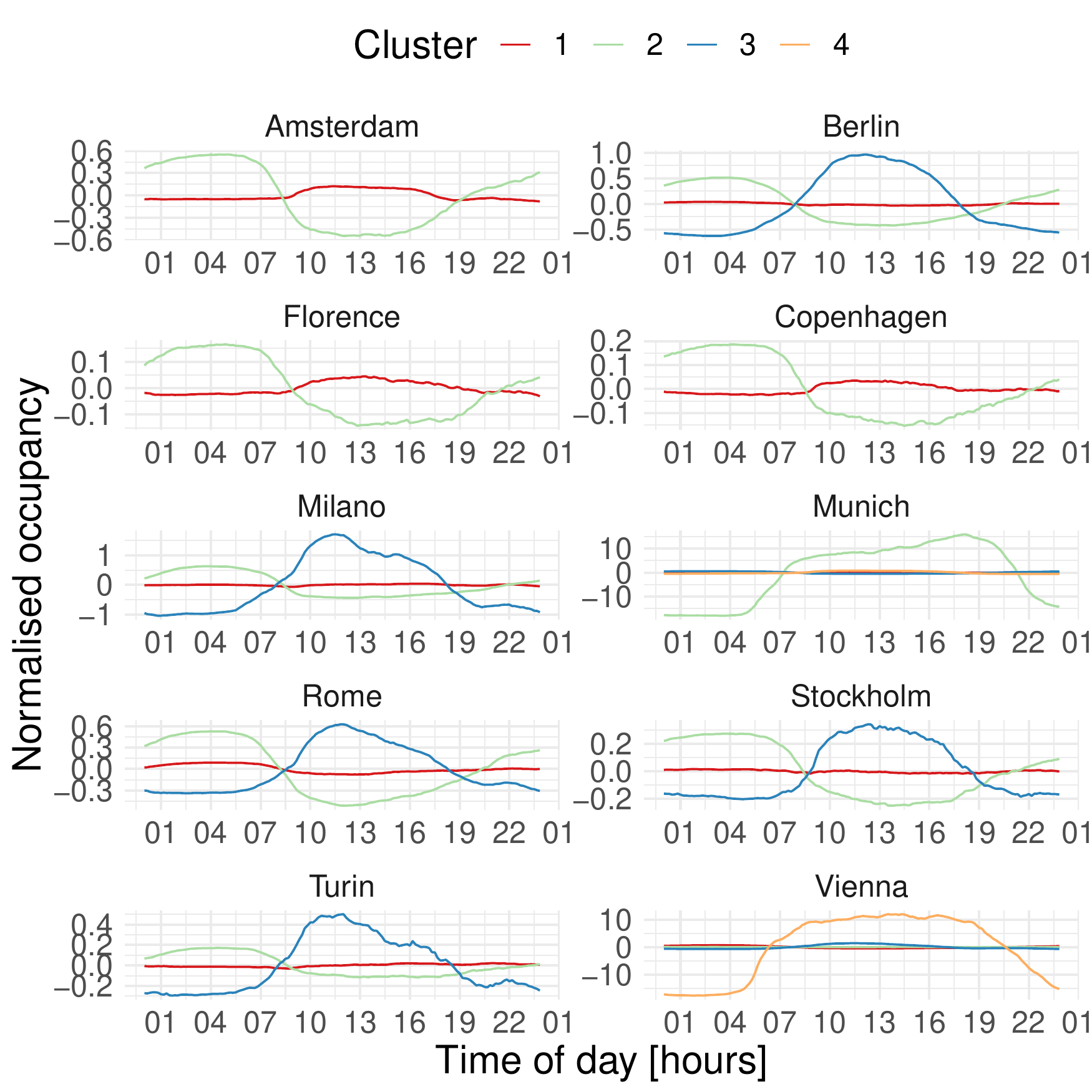}\fi
\caption{\csentence{Time series of vehicle availability per cluster in the ten cities.}}
\label{fig:clusters_ts}
\end{figure}

\begin{figure}[!h]
\centering
\iffigures\includegraphics[scale=0.45]{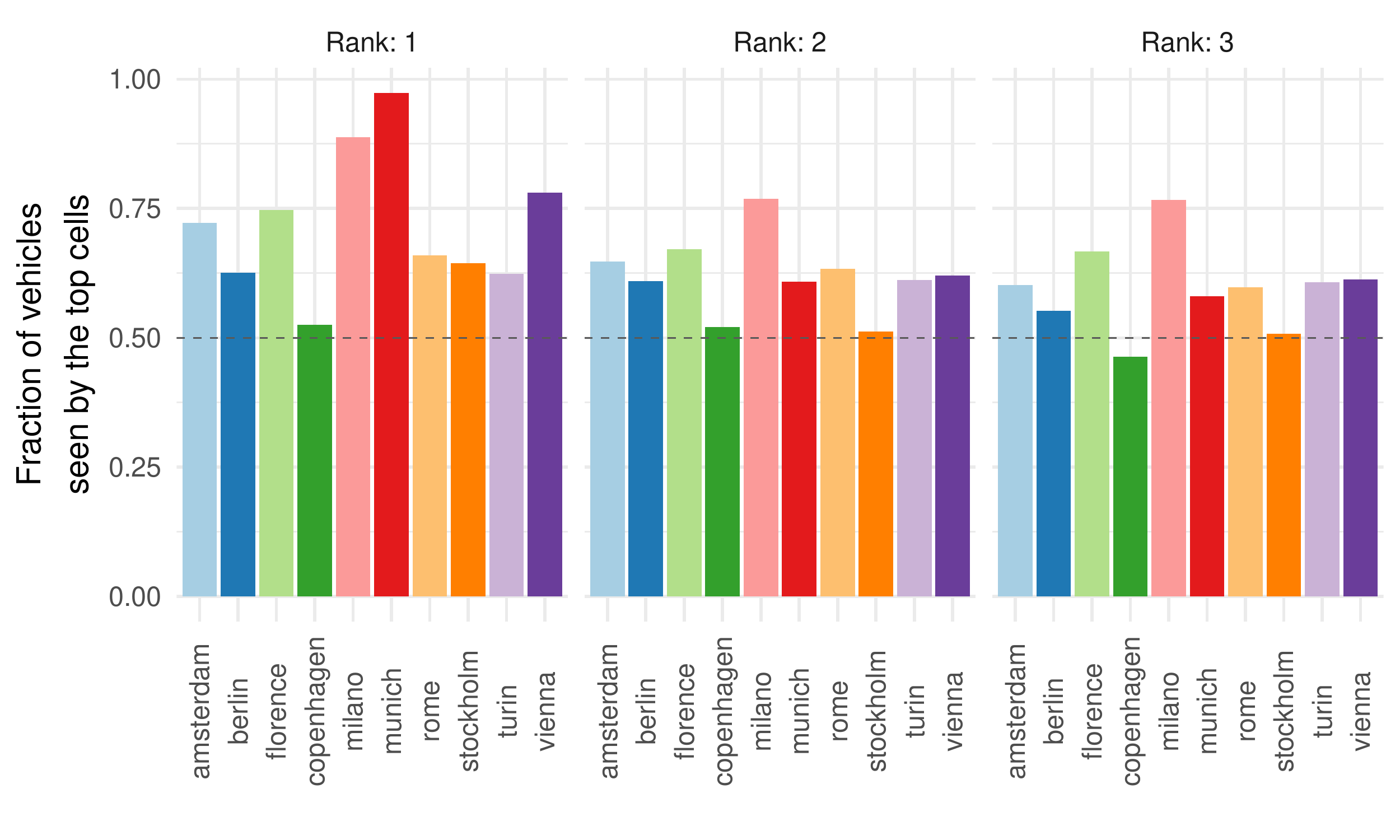}\fi
\caption{\csentence{Potential service areas. Tolerance window: 30 days.}}
\label{fig:service_areas_w30}
\end{figure}

\begin{figure}[!h]
\centering
\iffigures\includegraphics[scale=0.45]{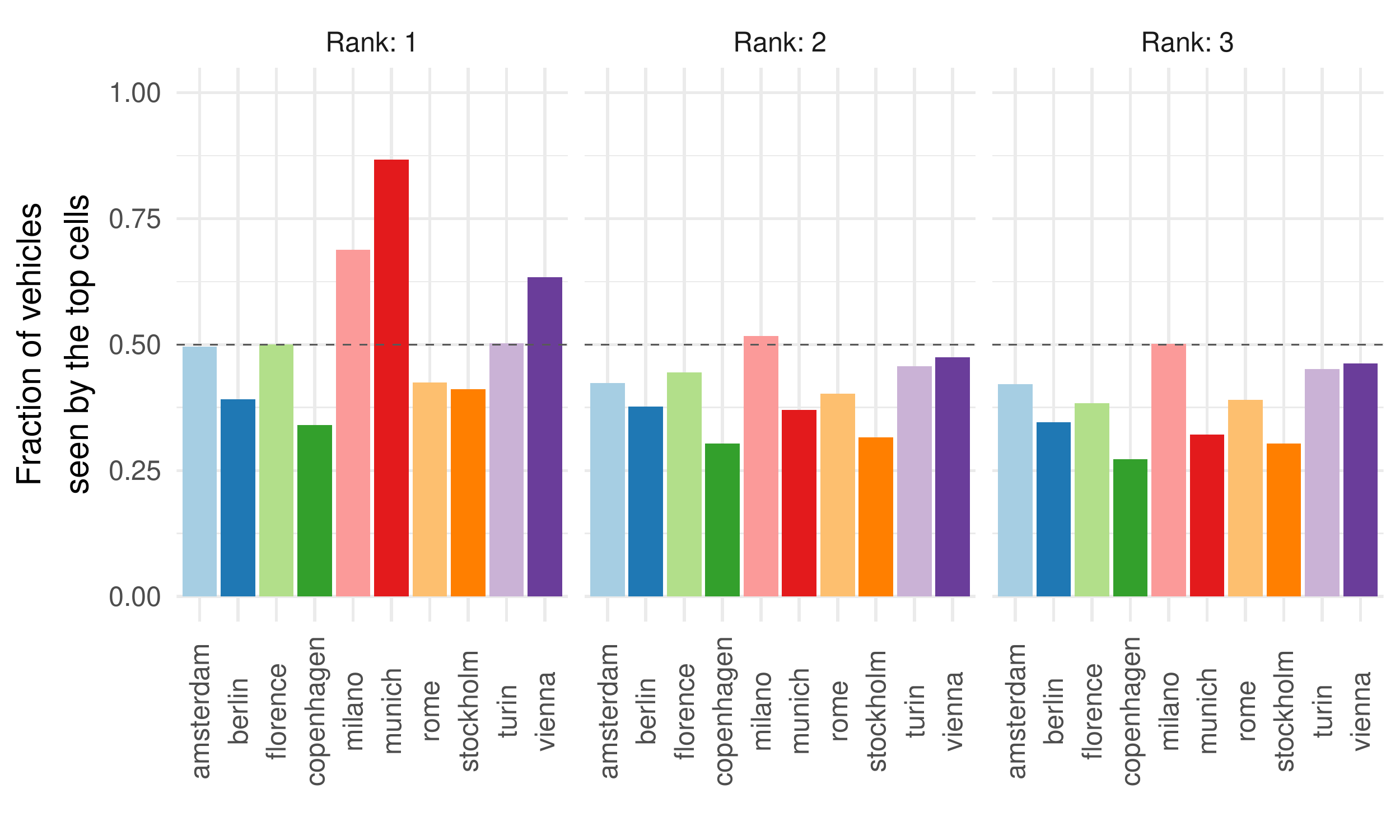}\fi
\caption{\csentence{Potential service areas. Tolerance window: 15 days.}}
\label{fig:service_areas_w15}
\end{figure}

%






%

\end{backmatter}
\end{document}